\documentclass[10pt,twocolumn,letterpaper]{article}

\usepackage{iccv}
\usepackage{times}
\usepackage{epsfig}
\usepackage{graphicx}
\usepackage{amsmath,accents}
\usepackage{amssymb}

\usepackage[inline]{enumitem}
\usepackage{xcolor}
\usepackage{multirow}
\usepackage{booktabs}
\usepackage{xcolor}

\usepackage[pagebackref=true,breaklinks=true,letterpaper=true,colorlinks,bookmarks=false]{hyperref}

\iccvfinalcopy 

\def\iccvPaperID{6811} 

\ificcvfinal\fi

\begin{document}

\title{Box-Aware Feature Enhancement for Single Object Tracking on Point Clouds}
	\author{Chaoda Zheng$^{1, \dagger}$, Xu Yan$^{1, \dagger}$, Jiantao Gao$^{2}$,  Weibing Zhao$^{1}$, 
	Wei Zhang$^{3}$,  Zhen Li$^{1,}$\thanks{{ Corresponding author: Zhen Li. $^\dagger$ Equal first authorship.}}~, Shuguang Cui$^{1}$ \\
	$^{1}$The Chinese University of Hong Kong (Shenzhen), Shenzhen Research Institute of Big Data, \\
	$^{2}$Research  Institute  of USV  Engineering, Shanghai University,  $^{3}$Baidu Inc\\
	{\tt\small	\{{chaodazheng@link.}, xuyan1@link., {lizhen@}\}cuhk.edu.cn}
}

\maketitle
\ificcvfinal\thispagestyle{empty}\fi

\begin{abstract}
   Current 3D single object tracking approaches track the target based on a feature comparison between the target template and the search area. 
   However, due to the common occlusion in LiDAR scans, it is non-trivial to conduct accurate feature comparisons on severe sparse and incomplete shapes.
   In this work, we exploit the ground truth bounding box given in the first frame as a strong cue to enhance the feature description of the target object, enabling a more accurate feature comparison in a simple yet effective way.
   In particular, we first propose the {\textbf{BoxCloud}}, an informative and robust representation, to depict an object using the point-to-box relation.
   We further design an efficient box-aware feature fusion module, which leverages the aforementioned BoxCloud for reliable feature matching and embedding.
   Integrating the proposed general components into an existing model P2B~\cite{qi2020p2b}, we construct a superior {\textbf{box-aware tracker (BAT)}}\footnote{~\url{https://github.com/Ghostish/BAT}}.
   Experiments confirm that our proposed BAT outperforms the previous state-of-the-art by a large margin on both KITTI and NuScenes benchmarks, achieving a {\textbf{15.2\%}} improvement in terms of precision while running $\sim$\textbf{20\%} faster.
\end{abstract}

\section{Introduction}\label{sec:intro}
Single object tracking (SOT) in 3D scene has a broad spectrum of practical applications, such as autonomous driving~\cite{luo2018fast}, semantic understanding~\cite{yan2020pointasnl,yan2020sparse} and assistive robotics~\cite{machida2012human,comport2004robust}. 
Given a 3D bounding box (BBox) of an object as the template in the first frame, the SOT task is to keep track of this object across all frames.
In real scenes, LiDAR becomes a popular 3D sensor due to its precise measurement, reasonable cost and insensitivity to ambient light variations. 
In this paper, we focus on SOT on LiDAR data, which can be viewed as 3D point clouds in general. 

\begin{figure}[t]
		
	\center\includegraphics[width=0.95\columnwidth]{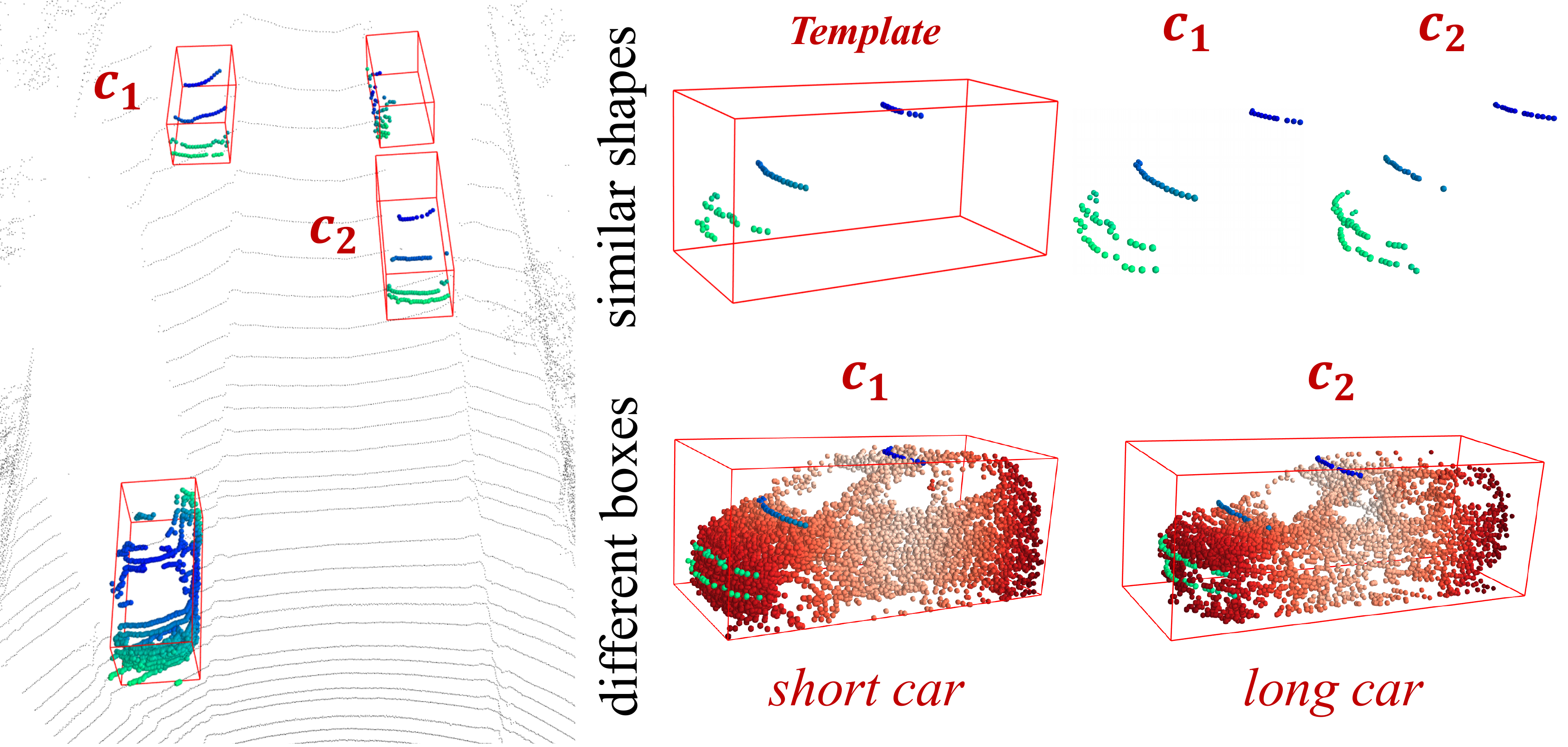}
		
	\caption{\textbf{Motivation behind BAT.} Partial scans of two different cars (\ie c1, c2) collected by a LiDAR sensor both have highly similar shapes with the target template (\textbf{1st row}), though their 3D BBoxes are quite different in size (\textbf{2nd row}). With explicit consideration of object bounding-boxes, our method can elegantly overcome such ambiguities. }
	\label{fig1}
	
\end{figure}

Due to the moving environment and self-occlusion, point clouds generated by a LiDAR system are inevitably irregular and incomplete, making the SOT task very challenging.
In 3D SOT, feature comparison plays an important role. The general idea to locate the target object is based on measuring the feature similarity between some candidate regions and the object template (initialized as the point cloud in the first given BBox).
For example, SC3D~\cite{Giancola_2019_CVPR} uses the exhaustive search or Kalman Filter to generate a set of candidate shapes at the current frame, and comparing them to the template using a siamese network. The candidate with the maximum similarity is chosen to be the target object for the frame.
Inspired by the success of the siamese region proposal network (RPN)~\cite{li2018high} in 2D SOT, P2B~\cite{qi2020p2b} proposes a point-based correlation based on the pair-wise feature comparison. P2B executes such a correlation between the template and the search area to output the target-specific search features, on which a 3D RPN is applied to obtain the final target proposals.
However, the features used for comparison are extracted from pure LiDAR point clouds, which face the following defects:
\begin{enumerate*}[label=\arabic*)]
  \item They do not encode the size information of objects. Since objects in LiDAR scans are mostly incomplete, it is hard to infer an object's size only from the partial point cloud. 
   \item They cannot capture the explicit part-aware structure information within each object BBox, \eg some part belongs to the car front while others belong to the sunroof as Figure~\ref{fig1} shows.
\end{enumerate*}
Therefore, the feature comparison among such features may bring considerable ambiguities which weaken the tracking performance.
What is a good representation for feature matching under 3D SOT? 
We revisit this problem by pointing out that the size and the part priors of the target object can be directly inferred from the template BBox given at the first frame. Based on this observation, we propose to address the above issues by explicitly utilizing the BBox to enhance the object features. 
Thus, we propose the BoxCloud, a robust and informative object representation depicting the point-to-box relation. Instead of using the $xyz$ coordinate, it represents an object point via a canonical $box~coordinate$, where the $i$-th dimension corresponds to the Euclidean distance between the object point and its $i$-th box point (\ie the corner or center of a BBox). 
Unlike the original LiDAR point cloud, a BoxCloud is defined based on both the object and its BBox.
Therefore, it naturally encodes the size and part information of an object. Note that we can directly compute the BoxCloud for a target template using the BBox given at the first frame. After the supervised training using ground-truth object BBoxes, the BoxClouds of objects in the search area can be easily predicted for the inference usage.

Based on the BoxCloud representation, the box-aware feature fusion (BAFF) module is further proposed to perform a correlation between the template and the search area to generate the target-specific search area features. It first measures the similarity between the search area and the template according to their BoxClouds. After such effective feature comparison, the BAFF module aggregates the top-k similar template points into each corresponding searching point, yielding a high-quality target-specific search area.
Finally, we construct a Box-Aware Tracker (BAT) by integrating the two proposed components into P2B~\cite{qi2020p2b}. By taking the auxiliary 3D BBox as input, BAT captures more shape constraints and part-aware information, no matter whether the input shape is partial or not, enabling effective and robust tracking on LiDAR point clouds.

Our main contributions can be summarized as follows:
\begin{itemize}
\setlength{\itemsep}{0pt}
\setlength{\parsep}{0pt}
\setlength{\parskip}{.5pt}
   \item To the best of our knowledge, we are the first to use free box information to boost the performance on the 3D SOT task. Specifically, we improve the feature comparison by designing a size-aware and part-aware BoxCloud feature, which is not only interpretable but also robust to sparseness and incompleteness.
   \item We propose a dedicated box-aware feature fusion module to generate better target-specific search areas in a box-aware manner.
   \item Experiments verify that our BAT achieves significant improvement over the state-of-the-arts on two benchmarks (\ie KITTI and NuScenes), especially on extremely sparse data.
\end{itemize}
\section{Related Work}
\noindent\textbf{2D Siamese Tracking.}
%
%
Recently, Siamese based methods~\cite{bertinetto2016fully,xu2020siamfc++,lukezic2020d3s,li2019siamrpn++,li2018high,zhu2018distractor,wang2019fast,zhang2019deeper,bhat2019learning} have demonstrated their advantages over those based on the traditional Discriminative Correlation Filter (DCF)~\cite{danelljan2016beyond,danelljan2017eco,bhat2018unveiling,xu2019joint,dai2019visual} in 2D visual object tracking. 
Siamese based trackers formulate the visual object tracking as a feature matching problem.
By utilizing the pre-trained feature extraction networks, siamese based trackers first project the target and search image onto a hidden embedding space and then compute their mutual similarity. 
%
%
Although 2D siamese based trackers have made remarkable progress in recent years, they focus on image patches and thus cannot be applied on point clouds directly.


\noindent\textbf{3D Single Object Tracking.}
%
%
Early 3D SOT methods~\cite{pieropan2015robust,bibi20163d,kart2018make,liu2018context,kart2019object} focus on the RGB-D information and tend to employ the 2D siamese architecture. 
SC3D~\cite{Giancola_2019_CVPR} proposes the first 3D siamese tracker on point clouds and introduces an auxiliary branch for shape completion to process the incomplete shapes. 
Although the object semantics can be learned via the shape completion task, SC3D lacks the ability of detecting the object relations in the search space. 
Moreover, it has low efficiency in target proposal generation and cannot be trained in an end-to-end manner. 
The follow-up~\cite{zarzar2019efficient} accelerates SC3D by leveraging the SiamRPN~\cite{li2018high} to generate target proposals from the bird eye views. However, its RPN only operates in 2D, and thus it cannot sense the 3D object relations.
To address these, P2B~\cite{qi2020p2b} adapts the SiamRPN~\cite{li2018high} to the 3D case where pure point clouds are processed.
Firstly, the target object information is fused into the search space. 
Following that, an object detection network is applied to the search space to detect the target.
Equipped with a state-of-the-art detection network, \ie VoteNet~\cite{qi2019deep}, P2B achieves significant improvement in terms of both accuracy and efficiency.
%
%
Very recently, Feng~\textit{et al.}~\cite{feng2020novel} presents a two-stage object re-track framework for 3D
point clouds, which can re-track the lost object at the coarse stage. 
However, none of these methods exploit the BBox provided in the first frame as an additional cue.
The information loss incurred by the incompleteness of shapes cannot be compensated, and their performance drops drastically when they deal with incomplete scans. 

\begin{figure*}[t]
	\center\includegraphics[width=\linewidth]{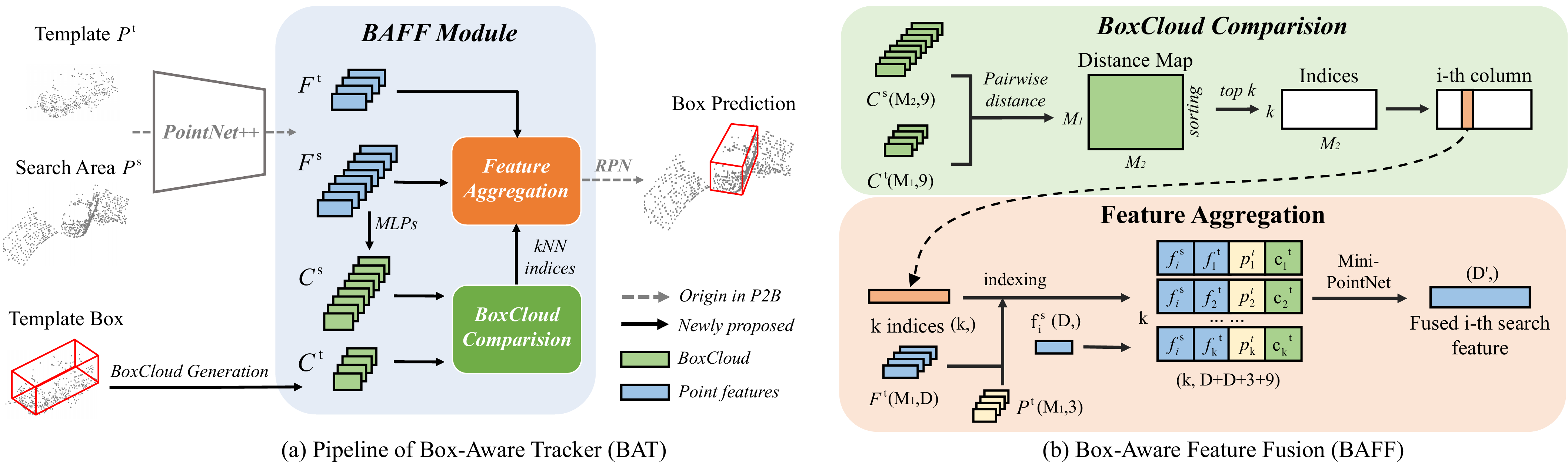}
	\caption{\textbf{Box-Aware Tracker (BAT) pipeline and Box-Aware Feature Fusion (BAFF) module.} 
	Part (a) illustrates the pipeline of BAT. 
    %
    The \textit{RPN} component denotes the 3D RPN used in \cite{qi2020p2b}, which is exploited to generate the final 3D target proposals.
    Part (b) illustrates the workflow of BAFF module. It first uses the distance map of BoxClouds as the metric to find out the k-nearest neighbors in the template point cloud with respect to each search area point. Then, a Mini-PointNet is used to aggregate the retrieved neighbors' features.
    }
	\label{fig2}
\end{figure*}
\noindent\textbf{3D Multi-Object Tracking.}
Most of the 3D Multi-Object Tracking (MOT) methods follow the tracking-by-detection paradigm~\cite{Weng2020_AB3DMOT,chiu2020probabilistic,yin2021center,zhang2019robust,patil2019h3d,wu20213d}.
Unlike SOT where the 3D BBox of the target is provided in the first frame, the MOT tracker determines the number of objects in all frames by running an independent detector~\cite{shi2019pointrcnn,shi2020points,shi2020pv}.
After that, the tracking is done by performing data association on the detection results (\ie linking the detected object BBoxes across all frames to obtain full trajectories), according to the motion estimation.
The motion of the objects can be estimated using the handcrafted Kalman filter~\cite{Weng2020_AB3DMOT,patil2019h3d}, or some learned features~\cite{zhang2019robust,yin2021center}.

%
\noindent\textbf{Bounding Box Utilization in 3D.}
In the 3D SOT task, though the BBox of the object is provided in the first frame, the previous methods~\cite{Giancola_2019_CVPR,qi2020p2b} only expand the BBox with a fixed magnification and crop out the point cloud inside as a template. 
In 3D detection, \cite{shi2020points} predicts intra-object part locations of all 3D points within a proposal, using the free-of-charge part supervisions derived from 3D ground-truth BBoxes. However, it only focuses on part-aware information while ignores the sizes of BBoxes. Moreover, it can only use BBoxes as supervision instead of the network input. 
In contrast, our method explicitly takes the given target BBox as input to enhance the target features with both part-aware and size-aware cues.

\section{Method}




\subsection{Problem Statement}
\label{ProblemDef}
A 3D BBox is represented as $(x,y,z,w,l,h,\theta) \in \mathbb{R}^7$, where $(x,y,z)$ stands for the coordinate of the BBox center, $(w,l,h)$ is the BBox size and $\theta$ is the heading angle (the rotation around the $up$-axis). A point cloud is represented as $\{p_i\}_{i=1}^{N}$, where each point $p_i$ is a vector of its $(x,y,z)$ coordinate and $N$ is the number of points.

In 3D single object tracking (SOT), the BBox of the target in the first frame is given to the tracker. The goal of the tracker is to locate the same target in the search area frame by frame. Notice that in the 3D case, any BBox is \textit{amodal} (covering the entire object even if only part of it is visible). A template point cloud $P^t = \{p^t_i\}_{i=1}^{N_t}$ is generated by cropping and centering the target in the first frame with the given BBox $B^t$. For any search area $P^s = \{p^s_i\}_{i=1}^{N_s}$ in other frames, previous trackers take in the pair $(P^t,P^s)$ and output the amodal BBox of the target in the search area. Formally, previous trackers can be formulated as:
\begin{align}
   \textit{track}: \mathbb{R}^{N_t \times 3} \times \mathbb{R}^{N_s \times 3} \rightarrow \mathbb{R}^4, \\
   \textit{track}(P^t,P^s) \mapsto (x,y,z,\theta). 
   \label{track}
\end{align}
Note that the output has only 4 elements since we do not need to re-predict the size of the 3D BBox, which is unchanged across all frames.

Since previous works compare the search area only with the template that is usually incomplete, they lack the ability to capture the size and part information of the target object. Therefore, they are prone to produce false target proposals when similar objects are present. In this work, we aim at designing a box-aware tracker $\textit{track}_\textit{box-aware}$ that exploits the amodal BBox of the target:
\begin{align}
   \textit{track}_\textit{box-aware}: \mathbb{R}^7 \times \mathbb{R}^{N_t \times 3} \times \mathbb{R}^{N_s \times 3} \rightarrow \mathbb{R}^4, \\
   \textit{track}_\textit{box-aware}(B^t,P^t,P^s) \mapsto (x,y,z,\theta). 
   \label{track_box_aware}
\end{align}

\subsection{Box-Aware Tracker (BAT)}
\label{BAT}
%
Our box-aware tracker is built upon P2B~\cite{qi2020p2b}. Following the siamese paradigm, P2B first generates a target-specific search area by executing a point-wise correlation between the template and the search features, which are extracted by a shared backbone. The correlation is based on the point-wise similarity and utilizes PointNet~\cite{qi2017pointnet} to achieve permutation invariance.
Then it applies a VoteNet~\cite{qi2019deep} based RPN to generate target proposals. Specifically, a voting module votes the seed points to potential target centers. Besides, a seed-wise targetness score is also predicted to regularize the feature learning. These potential target centers are then grouped into clusters using ball query~\cite{qi2017pointnet++}, which are finally turned into target proposals through another PointNet.
Each proposal is a 5D vector containing the coordinates of the target center (3D), the rotation in the horizontal plane (1D) and the targetness score (1D). The proposal with the highest targetness score is chosen as the final prediction.

Our work focus on improving the first part of P2B~\cite{qi2020p2b} because generating a better target-specific search area is critical for robust tracking.
The overall architecture of BAT is shown in Figure~\ref{fig2} (a). Given a template point clouds $P^t \in \mathbb{R}^{N_1\times 3}$ and a search area $P^s \in \mathbb{R}^{N_2\times 3}$, a shared PointNet++~\cite{qi2017pointnet++} is utilized to extract features from both of them, resulting in $F^t \in \mathbb{R}^{M_1\times D}$ and $F^s \in \mathbb{R}^{M_2\times D}$.
Besides, the BoxCloud $C^t \in \mathbb{R}^{M_1\times 9}$ (Section~\ref{sec:BoxCloud}) is obtained from the template and its BBox.  
The BoxCloud of search area $C^s$ is predicted by a multi-layer perceptron (MLP).
After that, through BoxCloud comparison and feature aggregation sub-modules (Section~\ref{sec:BAFF}), we obtain the target-specific search area $\hat F_s \in \mathbb{R}^{M_2\times D}$.
After getting $\hat F_s$, we yield the final target proposals using the same way as~\cite{qi2020p2b}.

\subsection{BoxCloud Representation}
\label{sec:BoxCloud}
%
%
Previous works only use the BBox given at the first frame to crop and center the target during the pre-processing, ignoring that it contains rich shape information about the target. For a BBox $(x,y,z,w,l,h,\theta)$ of an object, its $(w,l,h)$ indicates the size of the full object, even if part of the object is occluded or truncated. Its $(x,y,z,\theta)$ indicates the object coordinate system. Knowing the size and the object coordinate system, we can infer the intra-object location of each point, \ie the part information. 
Based on these observations, we design the BoxCloud representation to fully utilize such shape priors.

A BoxCloud is defined by the point-to-box relation between an object point cloud $P$ and its BBox $B$. For each point $p_i$ in $P$, we first calculate its Euclidean distance to each of the eight corners and the center of $B$, resulting in nine distances. Then we arrange them in a predefined order with respect to the object coordinate system, generating a 9-D vector $c_i$ (as shown in Figure~\ref{fig3}). We denote $c_i$ as the \textit{box coordinate} of $p_i$ with respect to $B$. A BoxCloud is a set of box coordinates and can be formulated as follows:
\begin{align}
   C = \{c_i \in \mathbb{R}^9~|~c_{ij} = ||p_i - q_j||_2, ~~\forall j \in [1,9] \}_{i=1}^{N},
\end{align}
where $q_{j (j\neq 9)}$  is the $j$-th corner and $q_9$ is the center of $B$.

%
Though simple, BoxCloud features own the following properties over the geometric point clouds (\ie spatial $xyz$ coordinates of point clouds):
\begin{figure}[t]
		
	\center\includegraphics[width=\linewidth]{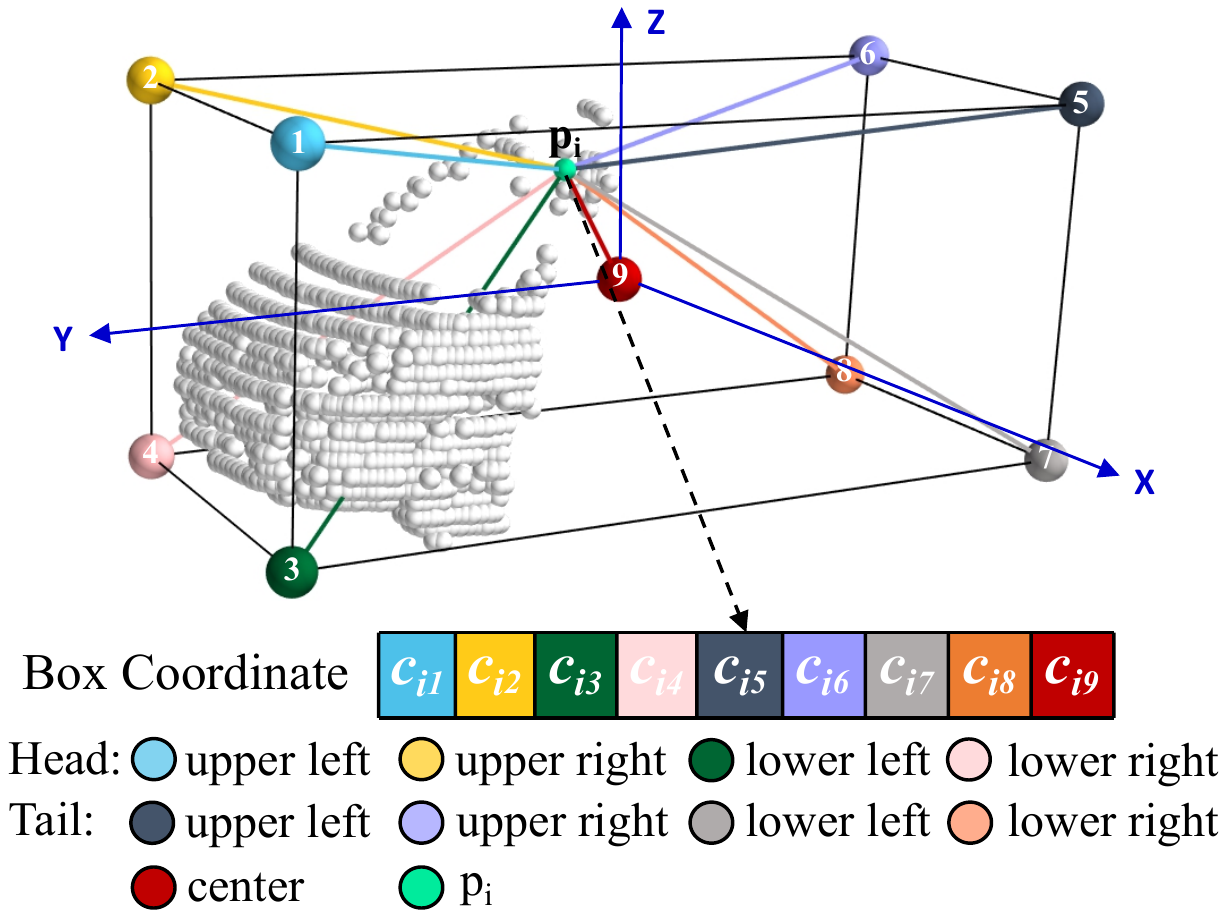}
		
	\caption{\textbf{BoxCloud representation.} BoxCloud depicts the distances between object points and box points (\ie corners and the center of a 3D BBox). Box points are arranged in a predefined order. The figure shows the $box~coordinate$ of an object point $p_i$. The object and its BBox are transformed to the object coordinate system with $x$-axis pointing to the left, $y$-axis pointing to the front and $z$-axis pointing upward.}
	\label{fig3}
		
\end{figure}

\noindent\textbf{Size-Awareness.} The full size of an object can be directly inferred from its BoxCloud regardless of shape incompleteness. This helps to distinguish similar object parts of incomplete scans from objects of different sizes.

\noindent\textbf{Part-Awareness.} Each $c_i$ indicates the intra-object location of $p_i$, because $c_i$ encodes the relative position of $p_i$ with respect to the corners and center. For example, if $\mathrm{argmin}_j~c_{ij} = 1$, which mean $p_i$ is closest to $q_1$, we know $p_i$ belongs to the upper left part of the object head. 



\subsection{Box-Aware Feature Fusion}
\label{sec:BAFF}
The goal of the Box-Aware Feature Fusion (BAFF) module is to generate an enhanced target-specific search area by augmenting the search area with the template, which enables us to locate the target in the search area using a 3D RPN. BAFF can also be regarded as performing a correlation on search area features using the template features as the kernel.
BAFF mainly consists of the \textit{BoxCloud Comparison} and the \textit{Feature Aggregation} sub-modules, which are described in details as below.

\subsubsection{BoxCloud Comparison}

BAFF depends on BoxCloud comparison to conduct a reliable comparison between the template and the search area. 
%
Considering the BoxCloud features of the search area are unknown during testing, we first utilize a point-wise MLP to predict the BoxClouds of objects in the search areas under the supervision of ground truth BBoxes.

Given the features of a search area $F^s =\{f_i^s \in \mathbb{R}^{d}\}_{i=1}^{M} $ extracted by the backbone network, the MLP takes each feature point $f_i^s$ as input, and outputs its 9-D box coordinate $c_i^s$. The predicted $c_i^s$ is explicitly supervised by a Huber loss (\ie smooth-L1 regression loss):
\begin{align}
\mathcal{L}_{bc} = \frac{1}{\sum_i m_i}\sum_i ||c^s_i - \hat{c}^s_i||\cdot m_i,
\label{1}
\end{align}
where $\{\hat{c^s_i}\}_{i=1}^{M} \in \hat C^s$ are ground truth \textit{box coordinates} from the BoxCloud of the search area pre-calculated before training. $\{m_i\}_{i=1}^{M}$ is a binary mask indicating whether the $i$-th point is inside an object BBox ($m_i=1$) or not ($m_i=0$).

As shown in Figure~\ref{fig2} (b), after obtaining the predicted BoxCloud $C^s$ of the search area, we can conduct BoxCloud comparison between the predicted $C^s$ of the search area and the BoxCloud $C^t$ of the template. The BoxCloud comparison is simply based on the pairwise $l_2$ distance:
\begin{equation}
   \textit{Dist} \in \mathbb{R}^{M_1\times M_2} = \textit{Pairwise}(C^t,C^s), 
   \label{bc}
\end{equation}
where $\textit{Pairwise}(\cdot,\cdot)$ denotes the pairwise $l_2$ distance between the two input BoxClouds. $M_1$ and $M_2$ are the numbers of points in $C^t$ and $C^s$ respectively. \textit{Dist} denotes a $M_1$ by $M_2$ distance map whose entity $Dist_{i,j}$ represents the distance between ${c_i^t}$ and ${c_j^s}$.

Compared with the feature comparison using the extracted features~\cite{qi2020p2b}, measuring the shape similarity using BoxCloud is not only more reliable and interpretable but also more efficient. 
This is because the dimension of a learned feature must be much higher than nine (the dimension of the BoxCloud point) to encode sufficient information. However, a higher dimension inevitably brings additional computational costs.

After that, BAT takes the resulting distance map as a guide to sift out the $k$ most similar template points for each search point. 
The k-NN grouping is illustrated in Figure~\ref{fig2}~(b). 
%
BAT selects the top-$k$ most similar template points for each search area point,
resulting in an indices matrix where the $i$-th column contains the indices of the k nearest neighbors of the $i$-th search point in terms of BoxCloud distance. 
The k-NN grouping helps to discard false matchings and avoid noise interference from all template points, promoting the subsequent feature aggregation. 
Besides, using a smaller $k$ helps to gain noticeable speed up, which provides an alternative option for performance and latency trade-off (see Section~\ref{sec:abaltion}).

%

%

	
	
		

\subsubsection{Feature Aggregation}\label{sec:feature_aggregation}

Feature aggregation aims to fuse the top-$k$ related template features into the search area.
For the $i$-th point $p_i^s$ in the search area, we use the $i$-th column in the indices matrix to select its $k$ nearest template points. A mini-PointNet~\cite{qi2017pointnet} is then applied to aggregate these $k$ pairs \{$[p^t_j;f^t_j;c^t_j;f^s_i] \in \mathbb{R}^{2D + 3 + 9}, \forall j = 1,\ldots,k$\}. Here $p^t_j \in \mathbb{R}^{3}, ~f^t_j\in \mathbb{R}^{D}, ~c^t_j \in \mathbb{R}^{9}$ denote the spatial coordinate, extracted feature and \textit{box coordinate} of the j-th template point, respectively. $f^s_i \in \mathbb{R}^{D}$ is the extracted feature of the corresponding search point. Formally, the mini-PointNet can be formulated as follows:
\begin{align}
\hat{f}^s_i = {\texttt{MaxPool}}(\{\texttt{MLP}([p^t_j;f^t_j;c^t_j;f^s_i])\}^k_{j=1}),
\label{mini_pointnet}
\end{align}
where $\texttt{MLP}(\cdot)$ is a multi-layer perceptron and $\texttt{MaxPool}(\cdot)$ is the max-pooling across points. Finally, the template features are effectively and reliably encoded in the newly acquired target-specific search area $\hat F^s = \{\hat f_i^s\}_{i=1}^{M_2}$.
\begin{table*}
	\renewcommand\tabcolsep{6pt} 
	\small
		\caption{Performance comparison among our BAT and the state-of-the-art methods on the KITTI (left) and NuScenes (right) dataset, where the instance number of each category is shown under category names.
		%
		%
		\textbf{Bold} denotes the best performance.}
		\begin{center}

		\begin{tabular}{l|c|ccccc|ccccc}
				\toprule[.05cm]
				\multirow{3}*{}
				&  Approach & \multicolumn{5}{c|}{{KITTI}}  & \multicolumn{5}{c}{{NuScenes}}\\ 
				& Category & Car   & Pedestrian & Van & Cyclist &Mean & Car & Truck & Trailer & Bus  & Mean \\
				& Frame Number &\textit{6424} &\textit{6088} &\textit{1248} &\textit{308} &\textit{14068} &\textit{64159} &\textit{13587} &\textit{3352} &\textit{2953} &\textit{84051} \\
				\hline
				\hline
				\multirow{4}*{Success}
				& SC3D~\cite{Giancola_2019_CVPR} &41.3 & 18.2 &40.4 &41.5 &31.2 & 22.31 & 30.67 & 35.28 & 29.35 & 24.43 \\
				& SC3D-RPN~\cite{zarzar2019efficient} &36.3 & 17.9 & - &\textbf{43.2} & - & - & - & - & - & - \\
				& P2B~\cite{qi2020p2b} &56.2 &28.7 &40.8 &32.1 &42.4 &38.81 &42.95 &48.96 &32.95 &39.68\\
				& BAT(Ours) &\textbf{65.4} &\textbf{45.7} &\textbf{52.4} &33.7 &\textbf{55.0} &\textbf{40.73} &\textbf{45.34} &\textbf{52.59} &\textbf{35.44} & \textbf{41.76}\\
				\hline
				\hline
				\multirow{4}*{Precision}
				& SC3D~\cite{Giancola_2019_CVPR} &57.9 & 37.8 &47.0 &70.4 &48.5 & 21.93 & 27.73 & 28.12 & 24.08 & 23.19 \\
				& SC3D-RPN~\cite{zarzar2019efficient} &51.0 & 47.8 &- &\textbf{81.2} &- & - & - & - & - & - \\
            & P2B~\cite{qi2020p2b} &72.8 &49.6 &48.4 &44.7 &60.0 &43.18 &41.59 &40.05 &27.41 &42.24\\
				& BAT (Ours) &\textbf{78.9} &\textbf{74.5} &\textbf{67.0} &45.4 &\textbf{75.2} &\textbf{43.29} &\textbf{42.58} &\textbf{44.89} &\textbf{28.01} &\textbf{42.70}\\
				\toprule[.05cm]

			\end{tabular}
		\end{center}
		
		\label{tab:category}
 	\end{table*}

\subsection{Implementation}

\label{train and test}
\noindent\textbf{Search Area Generation.} In practice, the object movement between two consecutive frames is relatively small and hence there is no need to search the target in the whole frame.  All we need is to search the target in the neighborhood of the previous object location. We follow this idea to generate the search areas for training and testing. During both training and testing, templates and their BBoxes are transformed to the object coordinate system before being sent to the model. 

\noindent\textbf{Training.} We generate training samples from any two consecutive frames at the time $t-1$ and $t$ and use them to train BAT in an end-to-end manner. The template is generated by merging the point clouds inside the first given BBox and $(t-1)$-th target BBox. 
We randomly shift the $(t)$-th target BBox, enlarger it by 2 meters in each direction and collect the points inside to generate the search area.
Our loss used for training the RPN is the same as that in \cite{qi2020p2b}, denoting it as $\mathcal{L}_{rpn}$, and the final loss for our model is $\mathcal{L} = \mathcal{L}_{bc} + \lambda \mathcal{L}_{rpn}$. We used grid search to tune the $\lambda$ and finally set it to 1. In fact, changing $\lambda$ does not affect the performance too much.
Our network is trained for 60 epochs using Adam optimizer with a batch size of 96. The learning rate of the network is initialized as 0.001 and is divided by 5 every 12 epochs. 

\noindent\textbf{Testing.} During testing, we track a given target across all frames in a sequence. This is achieved by applying the trained BAT frame by frame. For the current frame, we update the template by merging the point clouds in the first given BBox and the previous predicted BBox. W e enlarge the previous predicted BBox by 2 meters and collect the points inside to form the new search area.

\noindent\textbf{Architecture.} For the model architecture, $k$ is set as 4 in the BAFF module and the Mini-PointNet is constructed with a three-layer MLP, where the channel numbers of hidden layers are identically 256. 
We randomly sample $N_1 = 512$ points for each template point cloud $P^t$ and $N_2 = 1024$ points for each search area $P^s$.
They are then fed into a PointNet++ (as that in P2B~\cite{qi2020p2b}) to obtain corresponding features $F^t \in \mathbb{R}^{128\times 256}$ and $F^s \in \mathbb{R}^{64\times 256}$, which will be used for the subsequent processing.
All experiments are conducted using NVIDIA RTX-2080 GPUs. 

\section{Experiments}
In this section, we further present our experiment setting and superior results from different aspects.

\subsection{Experimental Setting}
We conduct extensive experiments on two widely-adopted datasets (\ie KITTI~\cite{Geiger2012CVPR} and NuScenes~\cite{caesar2020NuScenes}) to validate the effectiveness of the proposed BAT method. 
Both datasets contain point clouds scanned by LiDAR sensors. \\[.15cm]
\noindent\textbf{Datasets.} 
For KITTI dataset, it contains 21 outdoor scenes and 8 types of targets. 
The NuScenes dataset is more challenging, containing 1000 driving scenes across 23 object classes with annotated 3D BBoxes.
Furthermore, there are much more objects in a scene, and thus tracking on NuScenes dataset is more challenging since the target is easily submerged by other objects.
For KITTI dataset, we follow \cite{Giancola_2019_CVPR,qi2020p2b} to setup the traninig/valid/test splits for a fair comparison.
For NuScenes, we use the train\_track split of its training set for training and test on its validation set. We only consider the key frames during both training and testing. Tracklets where the first BBoxes contain no point are not considered during the evaluation. 
\\[.15cm]
\noindent\textbf{Metrics.} 
For evaluation metrics, we follow \cite{Giancola_2019_CVPR} to use One Pass Evaluation (OPE)~\cite{wu2013online} to measure Success and Precision. For a predicted BBox and ground-truth (GT) BBox, ``Success" is defined using the AUC for intersection over union between them. ``Precision" is defined as AUC for distance between two boxes’ centers from 0 to 2m.

\begin{figure}[t]

	\center\includegraphics[width=0.9\columnwidth]{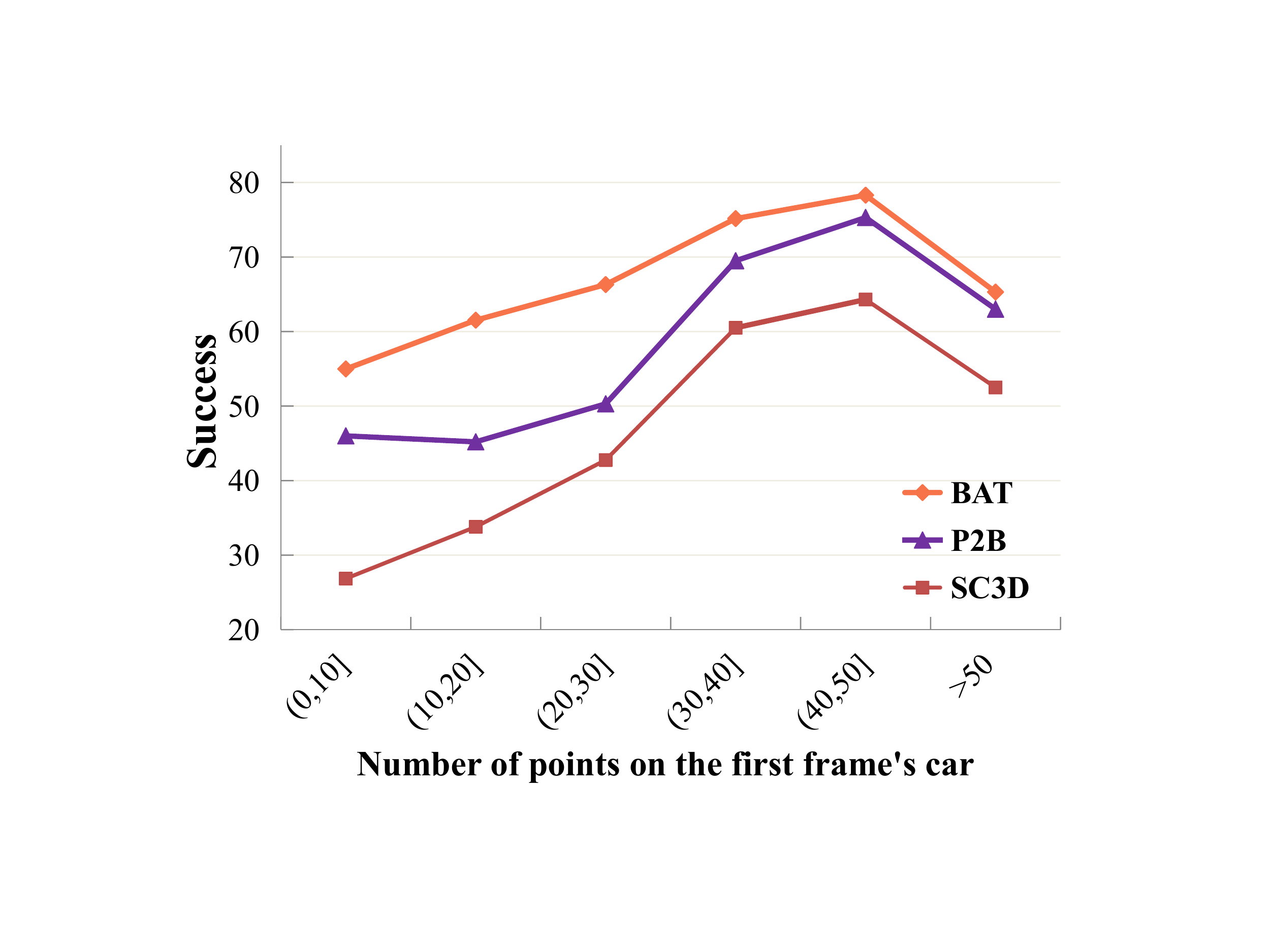}
	\vspace{.2cm}
	\caption{\textbf{Robustness test.} Methods are evaluated on sequences split by the number of points in the first frame’s car. We use the average Success for each interval (horizontal axis) as the evaluation metric.}
	\label{fig:sparsity}
		
\end{figure}

\begin{figure*}[t]
	\centering
	\center\includegraphics[width=0.95\linewidth]{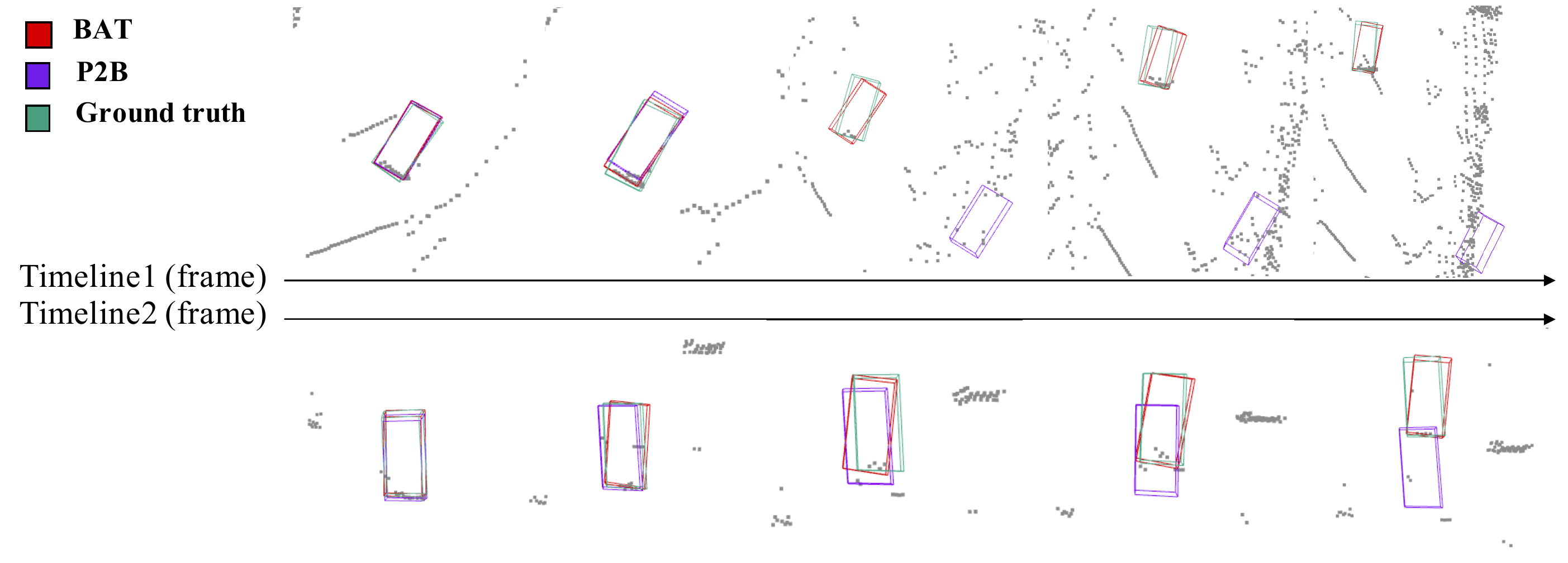}
	\caption{\textbf{Advantageous cases of our BAT compared with P2B}. We can observe BAT’s advantage over P2B in extremely sparse scenarios.}
	\label{fig:vis}
		
\end{figure*}

\begin{figure}[t]
	
\center\includegraphics[width=0.95\linewidth]{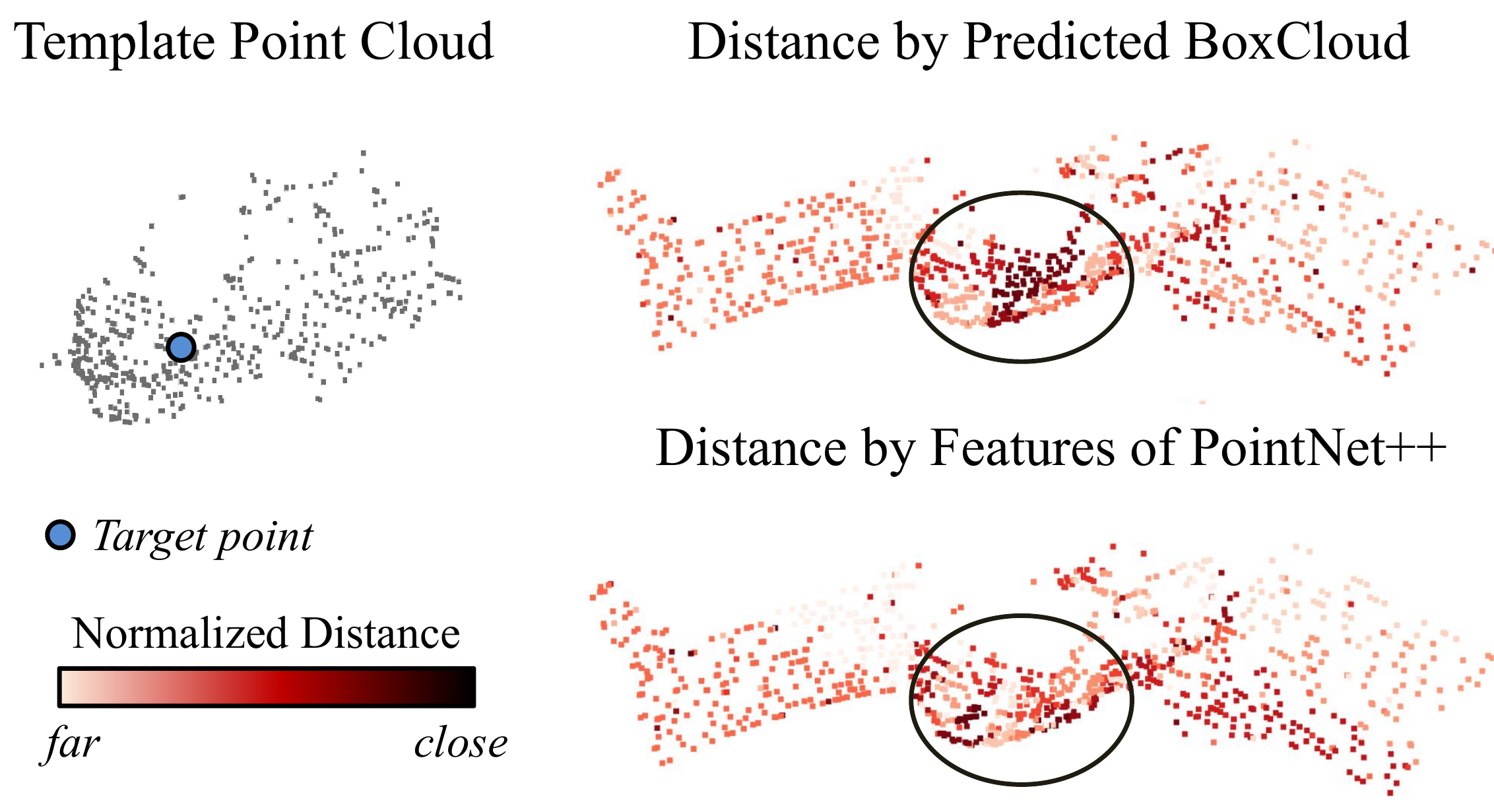}
	\vspace{.2cm}
\caption{\textbf{Pairwise distances measured on extracted features and BoxClouds.}
Distances are calculated between each point in the search area and the {\color{blue}{blue point}} in the template.}
\label{fig:vis_dis}
\end{figure}

\subsection{Comparison with State-of-the-arts}

We compare our method against SC3D~\cite{Giancola_2019_CVPR}, its follow-up~\cite{zarzar2019efficient} and P2B~\cite{qi2020p2b}. We do not include \cite{feng2020novel} in our main table because it does not provide open source codes and only has published results on car category of KITTI, which are 58.4/73.4 in terms of Success/Precision.

\noindent\textbf{Performance across Categories.} 
Table~\ref{tab:category} summarizes the results in the KITTI and NuScenes dataset. 
%
For the KITTI dataset, we report the published results from corresponding papers. For the NuScenes dataset, since there are no published results, we evaluate~\cite{Giancola_2019_CVPR,qi2020p2b} using their official open-source codes.
%

The left part of the table records the results on the KITTI dataset, where BAT shows a significant performance gain over existing methods, outperforming P2B by over 10\% on average. 
For the pedestrian category, BAT even achieves about 25\% improvement over P2B, which strongly demonstrates the effectiveness of our box-aware tracking pipeline. 
Moreover, for the van category with only 1248 instances, BAT still achieves satisfactory results, outperforming the other two methods by a large margin. This implies that the additional box information helps to reduce the demand for training data. 
For the category of cyclist, which has an extremely small amount of training samples, our results still out perform P2B notably.
%
Furthermore, as shown in the right part of Table~\ref{tab:category}, BAT is consistently superior to the other two competitors on four main classes in NuScenes, which is a more challenging dataset with more instances and object-interference.\\
\noindent\textbf{Robustness to Sparsity.} 
In realistic applications, objects collected by LiDAR sensors are mostly sparse and incomplete. According to \cite{qi2020p2b}, about 34\% of cars in KITTI hold fewer than 50 points. Therefore, the robustness against sparsity is an indispensable property for practical trackers. Figure~\ref{fig:sparsity} showcases the impressive robustness of BAT. 
Our method achieves over 55\% success rate even when dealing with targets holding less than 10 points. This result is 10\% higher than P2B and 30\% higher than SC3D.
For targets with fewer than 30 points, BAT still holds a huge performance advance over the previous SOTA. 
Even the performance gap narrows as the number of points approaches 50, our BAT still outperforms P2B by over about 2\% success rate.
This figure verifies that our BAT works not only for sparse point clouds, but also for dense ones.
We observe that the performance of all methods drops with $> 50$ points, which is counter-intuitive. To clarify this, we show that the average length for sequences with $>$50 points is much longer than that of sequences with $\le$50 points (86.3 vs 47.7), making the task much more difficult regardless of more points. The performance drop is mainly due to the accumulated errors when tracking on longer sequences.
Figure~\ref{fig:vis} visualizes BAT's advantage over P2B in extremely sparse scenarios, where BAT's predictions hold tight to the ground truth boxes when P2B tracks off course or even fails.

\noindent\textbf{BoxCloud Comparison v.s. Feature Comparison.}
Our proposed BoxCloud is a more effective representation compared with features extracted by neural networks.
Figure~\ref{fig:vis_dis} shows a case using PointNet++ to extract the features of the template and search point clouds.
It shows that the target points in the search area have relatively low similarities in this case, while the similarities calculated using BoxClouds are much more satisfying.
%
This result strongly proves that BoxCloud is much more reliable for feature comparison.

\subsection{Ablation Study}
\label{sec:abaltion}
In this section, comprehensive experiments are conducted to validate the design of the BoxCloud and the BAFF module. Behavior analysis of the proposed method is presented as well.
All ablated experiments are conducted on the car class of the KITTI dataset as P2B~\cite{qi2020p2b}.

Our box-aware feature module is specially designed for BoxCloud processing. 
However, since the BoxCloud provides additional clue for incomplete shapes, directly feeding it to the network must help to improve the performance. 
To validate this hypothesis, we add an extra branch to the pipeline of P2B, constructing a simple extension called \textbf{BAT-Vanilla}. 
It predicts the BoxCloud of the search area points. The BoxCloud of the template is also concatenated to the spatial coordinates for the feature fusion, while the other parts of P2B remain unchanged.
Results in the bottom part of Table~\ref{tab:ablation} confirm that the BoxCloud does help to improve the tracking performance. Specifically, BAT-Vanilla defeats P2B by a noticeable margin. 
Besides, the last two rows of Table~\ref{tab:ablation} prove the effectiveness of our BAFF module, implying that the BoxCloud comparison and kNN-grouping techniques help to further exploit the useful clues encoded in the BoxClouds.

\begin{figure}[t]

	\center\includegraphics[width=0.95\columnwidth]{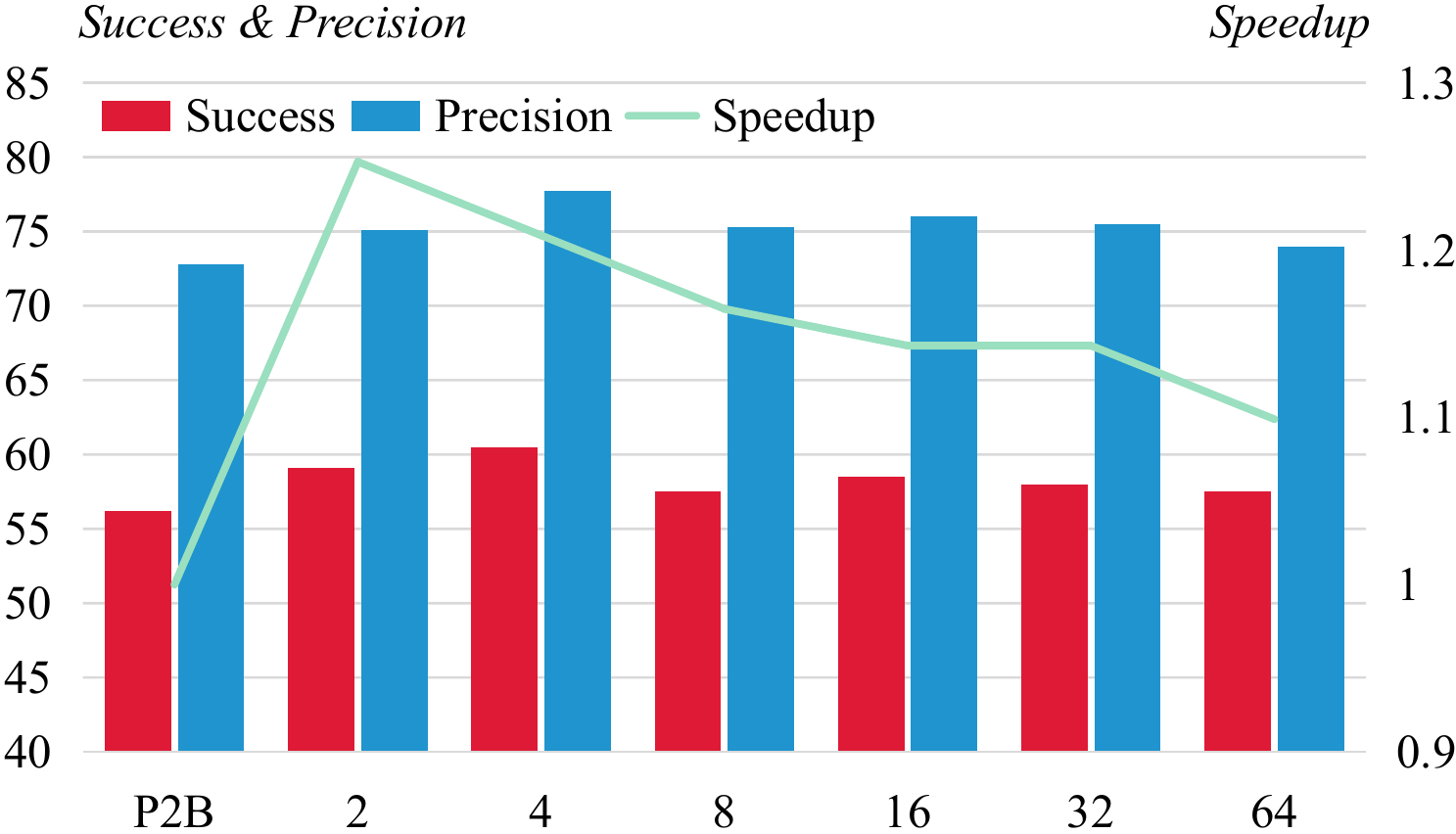}
	\caption{\textbf{Performance for different k values.} The left y-axis shows success and precision, while the right illustrates the speedup with respect to P2B.} 
	\label{fig:abla_k}
\end{figure}

\subsubsection{Analysis on Box-Aware Feature Fusion (BAFF)}

\noindent\textbf{The Choice of k}. The value of $k$ is a hyper-parameter of BAFF. 
Figure~\ref{fig:abla_k} shows that a smaller $k$ (\ie $k=2$) helps to speed up our model with little negative impact on the performance.
%
Furthermore, setting a larger $k$ does not improve the performance, which implies that the k-NN grouping in BAFF helps to reject false matching points in the template.
With the default setting $k=4$, BAT achieves the best performance with a satisfactory high speedup, about 20\% faster than P2B. 
 Also, it should be noted that BAT is stably superior to P2B under different $k$ values in terms of both performance and speed. Even with $k=64$ (\ie using all the template points same with P2B), it is still about 10\% faster than P2B. 
Note that here we only consider the time of model forwarding with batch size 96 when calculating the speedup. 
We do not consider the pre- and post-processing time because these operations are not optimized on GPUs.
During the inference where sequences are processed frame by frame, our BAT processes one frame with only 17.5ms on average, achieving 57 FPS.

\noindent\textbf{The Choice of Feature Aggregation}. In Eqn.~\eqref{mini_pointnet} of BAT, the feature we send for aggregation is $[p^t_j;f^t_j;c^t_j;f^s_i]$. In the upper part of Table~\ref{tab:ablation}, other settings for the feature aggregation are listed.
Firstly, without concatenating the template BoxCloud, our model suffers from a great performance decrease since it does not introduce useful clues to enhance the feature fusion.
Furthermore, we observe that adding additional information does not bring any improvement. 
Specifically, the results of adding the predicted BoxCloud of the search area, pairwise distances in Eqn.~\eqref{bc} and coordinates of the search point cloud all show drastic drops.
A possible explanation for this is that such features bring vast redundancies, which may confuse the feature fusion.

\noindent\textbf{Effectiveness of BoxCloud Comparison}. To further illustrate the effectiveness of the BoxCloud comparison, two settings are used to validate as follows:
\begin{enumerate*}[label=\arabic*)]
   \item \textit{BAT without BoxCloud Comparison}: We replace BoxCloud with feature (\ie the feature extracted by PointNet++) to compute the distance map.
   \item \textit{P2B with BoxCloud Comparison}: Keeping the BoxCloud comparison, we follow the way in P2B~\cite{qi2020p2b} for subsequent feature aggregation.
\end{enumerate*}
As shown in the middle part of Table~\ref{tab:ablation}, when using features of PointNet++ in comparison, the performance of BAT drops significantly, which demonstrates that the BoxCloud Comparison is the key of BAFF. 
While the result of P2B with BoxCloud Comparison implies that a different feature aggregation styles can also benefit from the BoxCloud comparison. 
The overall results show that the BoxCloud comparison and the BAFF module can benefit from each others. 


\begin{table}[]
\renewcommand\tabcolsep{3pt} 
\centering
\caption{\textbf{Results of different ablations.} The upper part shows different concatenation strategies for the feature aggregation sub-module (FA) and the middle part illustrates the effectiveness of the BoxCloud Comparison. As for the lower part, it shows our baseline (P2B), BAT-Vanilla (in Section~\ref{sec:abaltion}) and the best results.}
\begin{tabular}{l|cccc|cc}
 \toprule[1pt]
method & Success & Precision \\
\hline
\hline
FA without $C^t$ & 58.9 & 75.3\\
FA with additional $C^s$ & 56.7 & 74.7 \\
FA with additional $Dist$ & 58.4 & 75.4 \\ 
FA with additional $P^s$ & 56.7& 73.4 \\    
\hline
BAT without BoxCloud Comparison & 57.1  & 72.9 \\
P2B with BoxCloud Comparison& 58.3 &74.1  \\\hline
P2B (Baseline) & 56.2 & 72.8\\
BAT-Vanilla (P2B + BoxCloud) & 59.0  & 74.6 \\
BAT (Ours) & \textbf{60.5}  & \textbf{77.7}\\

\toprule[1pt]

\end{tabular}%
\label{tab:ablation}%
\end{table}%



   

\section{Conclusion}
In this paper, we enhance the SOT on LiDAR point clouds by fully exploiting the free bounding-box information.
Two main components of our work, the BoxCloud representation and the box-aware feature fusion module, are designed specially for effective object representation and box-aware target-specific search area generation. Each of them independently improves the tracking performance while benefits from each other when they are applied jointly.
We extend P2B with the two proposed components, constructing BAT. With such small modifications over P2B, our BAT outperforms the current state-of-the-art methods by a large margin, especially on sparser object tracking.
We believe that BoxCloud provides a flexible and powerful tool for future work to enhance their performance.
   %

\section*{Acknowledgment}
	
	{\noindent This work was supported in part by NSFC-Youth  61902335,  by Key Area R\&D Program of Guangdong Province with grant No.2018B030338001, by the National Key R\&D Program of China with grant No.2018YFB1800800,  by Shenzhen Outstanding Talents Training Fund, by Guangdong Research  Project No.2017ZT07X152, by Guangdong Regional Joint Fund-Key Projects 2019B1515120039,  by the  NSFC 61931024\&81922046, by helixon biotechnology company Fund and CCF-Tencent Open Fund.}

{
\bibliographystyle{ieee_fullname}
\bibliography{ref}
}

\newpage

	\begin{center}
		{\textit{\Large\bf Supplementary Material}}
	\end{center}

	\thispagestyle{empty}
	\setcounter{section}{0}
	\setcounter{figure}{0}
	\setcounter{table}{0}
	\renewcommand\thesection{\Alph{section}}
	
\section{Overview}
In this supplementary material, we provide more analysis experiments in Section~\ref{sec:exp}. Then we describe the architecture of BAT-Vanilla in Section~\ref{sec:vanilla}. 

\section{More Analysis Experiments}\label{sec:exp}
\noindent\textbf{Other BoxCloud Design}.
Our BoxCloud depicts the point-to-box relation using Euclidean distance. An alternative design is replacing the Euclidean distance with the \textit{point-to-point offset}. 
We denote BoxCloud in this form as $\accentset{\circ} B \in \mathbb{R}^{N\times 27}$, where each row of $\hat B$ is a concatenation of 9 three-dimensional offsets. 
Compared to our original design, $\accentset{\circ} B$ is three times larger but contains almost the same amount of information as ours.
%
%
As shown in Table~\ref{tab:bb_design}, using $\accentset{\circ} C$ incurs a performance decline,
which implies that the extra dimension of $\accentset{\circ} C$ puts a burden on the training of the network.
In contrast, the proposed BoxCloud is more compact and effective.

\begin{table}[h]
	\centering
	\caption{\textbf{Ablation on BoxCloud Designs.}}
	  \begin{tabular}{l|cc}
	  \toprule[1pt]
	   BoxCloud Designs & Success & Precision \\\hline \hline 
	  $\accentset{\circ} C$ (offset) & 58.8  & 75.3 \\
	  $C$ (original) & \textbf{60.5}  & \textbf{77.7} \\
	  \toprule[1pt]
	  \end{tabular}%
	\label{tab:bb_design}%
  \end{table}%
  
\noindent\textbf{Performance on Long/Short Term Tracking.} 
We further follow different schemas to generate search areas. 
1) To test the long-term tracking performance, we generate all the search areas based on previous results predicted by the models. 
In this setup, the trackers' ability to handle error accumulation and recover from failure is assessed. 
2) For short-term performance evaluation, we use the ground-truth location of the target in the previous frame to generate the next search area. In this case, the tracker does not have to handle the error introduced by its last prediction and only need to focus on the ``on time tracking" task.

\begin{table}[h]
			\small
				\caption{\textbf{Comparison on long/short term tracking performance for car}. The right two columns differ in their ways to generate search area. 
				\textbf{Bold} denotes the best performance.}
				\begin{center}
		
				\begin{tabular}{lcccc}
						\toprule[1pt]
						
								& Method & Long Term & Short Term\\
						\hline
						\hline
						\multirow{3}*{Success}
						& SC3D~\cite{Giancola_2019_CVPR} &41.3 & 64.6 \\
						& P2B~\cite{qi2020p2b} &56.2 &82.4 \\
						& BAT(Ours) &\textbf{60.5} &\textbf{83.5}  \\
						\hline
						\hline
						\multirow{3}*{Precision}
						& SC3D~\cite{Giancola_2019_CVPR} &57.9 & 74.5 \\
						& P2B~\cite{qi2020p2b} &72.8 &90.1 \\
						& BAT(Ours) &\textbf{77.7} &\textbf{90.5} \\
						\toprule[1pt]
		
					\end{tabular}
				\end{center}
				
				\label{tab:long_short}
		   \end{table}

The results of the car category for all the competitors are shown in Table~\ref{tab:long_short}. Overall, our BAT outperforms P2B and SC3D in both setups. 
In particular, BAT shows more notable superiority in long term setup.
%
%
This implies that the performance of our method is more stable and robust across time, while the other two methods are more likely to suffer from tracking failure. In the realistic tracking scenario, it is impossible to obtain the ``previous ground-truth".
Hence, it is more proper to use long-term performance to evaluate a practical tracking system.

 \begin{table}[h]
	\renewcommand\tabcolsep{4pt} 
	\small
	   \caption{\textbf{Different ways for template generation}. Methods are compared on the Car category. “First $\&$ Previous” denotes “The first GT and Previous result”.  
	   \textbf{Bold} denotes the best performance, and \underline{underline} shows our default setting.}
	   \begin{center}
 
	   \begin{tabular}{l|ccc|ccc}
			 \toprule[1pt]
			 
			 \multirow{2}{2.5cm}{\centering {Source of template}} & \multicolumn{3}{c|}{{Success}} & \multicolumn{3}{c}{{Precision}}\\
			 &BAT & P2B &SC3D &BAT & P2B &SC3D\\
			 \hline
			 \hline
			 The First GT &\textbf{51.8} &46.7 &31.6 
			 &\textbf{65.5} &59.7 &44.4\\
			 Previous result &\textbf{59.2} &53.1 &25.7 
			 &\textbf{75.6} &68.9 &35.1\\
			 \underline{First $\&$  Previous} &\textbf{60.5} &56.2 &34.9
			 &\textbf{77.7} &72.8 &49.8\\
			 All previous results &\textbf{55.8} &51.4 &41.3
			 &\textbf{71.4} &66.8 &57.9\\ 
			 \toprule[1pt]
 
		  \end{tabular}
	   \end{center}
	   
	   \label{tab:templete_gen}
	\end{table}

\begin{figure*}[t]
	\center\includegraphics[width=0.9\linewidth]{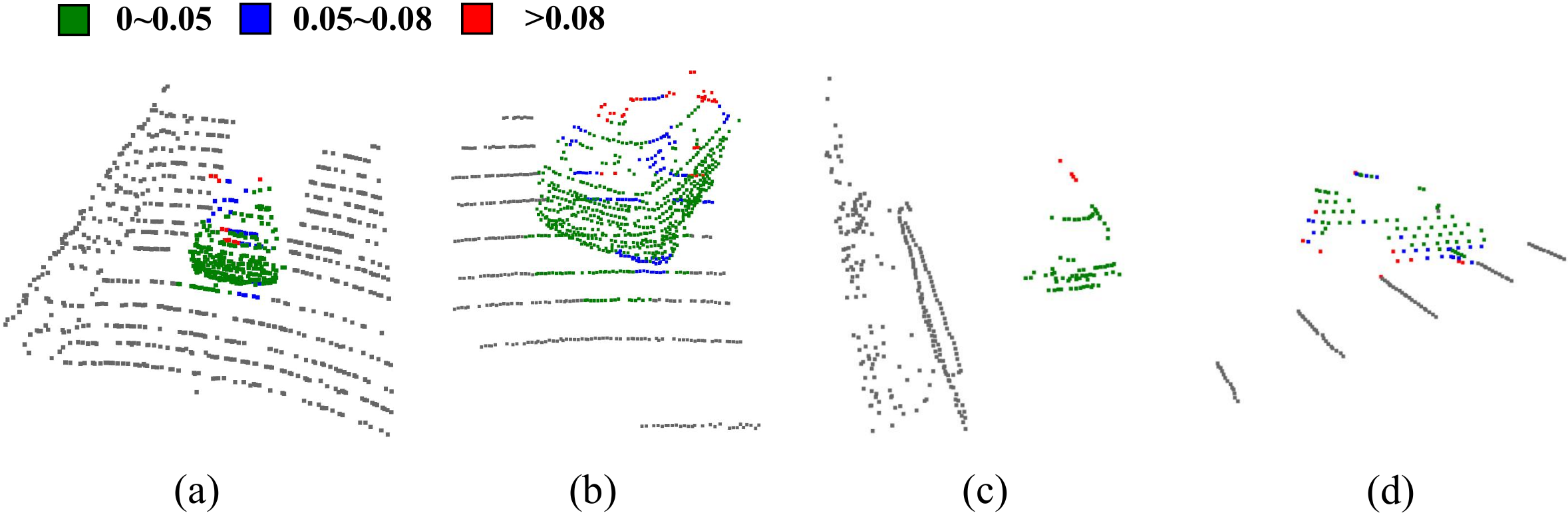}
	\caption{
    \textbf{Visualization of BoxCloud predictions.} 4 {\textit{car}} cases with different sparsity are presented. Points are colored according to its corresponding BoxCloud predictions MSE errors. The greens are points with MSE errors less than 0.05; The blues are points with MSE errors in the range between 0.05 and 0.08; while the reds denotes points with MSE errors higher than 0.08. It is obvious that most predictions are with small MSE errors.
    }
	\label{boxcloud_learning_visual}
\end{figure*}

\noindent\textbf{Template Generation Strategy.} During the testing, our default setting for template generation is to merge the target in the first frame (the ground truth) with the previous result predicted by the network.
For consistent comparison, we further test our method under another three template generation strategies, which uses ``the first ground-truth", ``the previous result", and ``all previous results" respectively to generate the template. The results are listed in Table~\ref{tab:templete_gen}. BAT maintains notable advantages regardless of any strategies. It is worth mentioning that our BAT defeats P2B by the largest margin ($\sim$7\%) under the ``previous result" strategy. This also shows that our long-term performance is much better and robust than that of P2B. 

 	\begin{table}[h]
		\small
		\caption{\textbf{Comparison with MOT methods on KITTI dataset.}}
		\begin{center}
			
			\begin{tabular}{lc|cccc}
				\hline
				Class & Method & AB3D~\cite{Weng2020_AB3DMOT} & PC3T~\cite{wu20213d} & BAT\\\hline
				Car & Succ/Prec & 37.5 / 42.3 & 51.9 / 59.2 & \textbf{60.5 / 77.7} \\
				Ped. & Succ/Prec & 17.6 / 27.3 & 23.6 / 34.1 & \textbf{42.1 / 70.1} \\
				\hline
				
			\end{tabular}
		\end{center}
		
		\label{tab:mot}
	\end{table}

	\noindent\textbf{Comparison with MOT Approaches.} To illustrate the superiority of single object tracking (SOT) upon 3D multi-object tracking (MOT) methods, we evaluate  3D MOT methods with 3D SOT metrics. 
	Since multiple objects need to be tracked in 3D MOT, we first find out the corresponding relation between the multiple objects with the single one.
	Specifically, for each object in SOT, we search its nearest neighbor in all objects of MOT.
	Hence, each MOT task can be transferred into several SOT tasks, and the metrics of SOT can play a normal role.
	For each object in MOT, if its tracked identity is changed in a specific frame (the tracker infer the error relation), we will stop the further tracking.
	
	We compare the two most representative methods in 3D MOT, AB3DMOT~\cite{Weng2020_AB3DMOT} and PC3T~\cite{wu20213d}, where they rank 25 and 4 in the KITTI MOT leaderboard in the car category, respectively.
	As shown in Table~\ref{tab:mot}, BAT significantly performs better than state-of-the-art 3D MOT methods.
	Since multiple objects need to be tracked, they cannot use target-specific feature augmentation to enhance the template representation.
	Moreover, since they use detection to obtain all objects in the scene, their speed is much slower than ours, where both two MOT methods cannot achieve real-time speed.

\noindent{\textbf{Visualization of BoxCloud Learning.}}\label{sec:bc_learning}
BAT is trained to predict the BoxClouds of the points in search areas. In this part, we compare the BoxClouds predicted by our trained BAT with the ground-truths.
\begin{figure}[h]
	\center\includegraphics[width=0.95\linewidth]{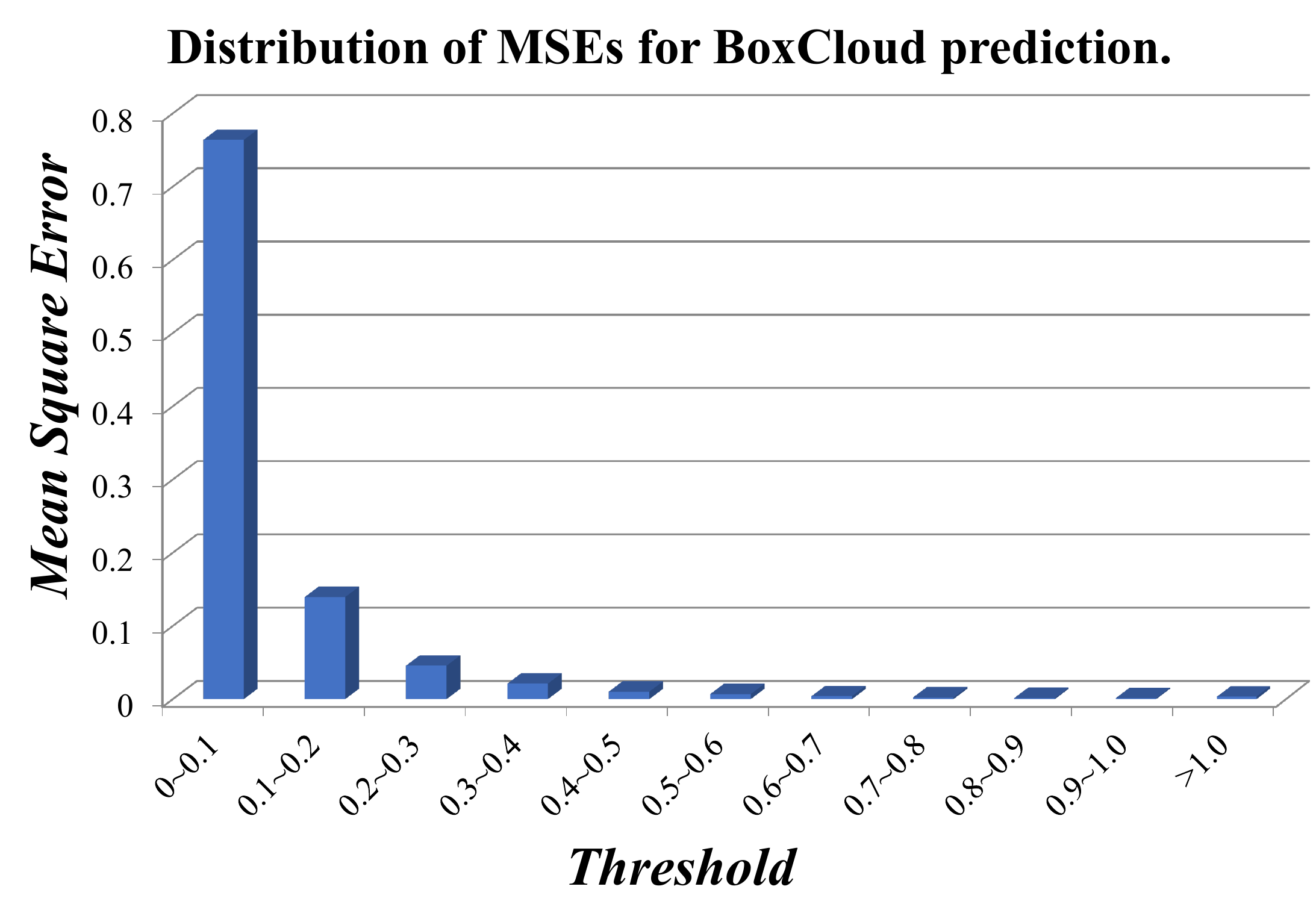}
	\caption{
    Distribution of MSEs for BoxCloud prediction.
    }
	\label{boxcloud_learning}
\end{figure}
We use the Mean Square Error (MSE) as the metric to evaluate the performance of BoxCloud learning. For each predicted BoxCloud point, we calculate the MSE between the corresponding ground-truth and the predicted BoxCloud point (only the target points are considered). Figure~\ref{boxcloud_learning} shows the distribution of MSEs of all the predictions in our KIITI test split. Most of the MSEs are less than 0.1, which implies a high accuracy of the BoxCloud prediction. 

We further visualize several cases of BoxCloud predictions in Figure~\ref{boxcloud_learning_visual}.
As shown in figures, our BAT can generally obtain accurate BoxCloud predictions. Some biases may occur in the edges between objects and backgrounds. This is because our training strategy only supervises the BoxCloud of the object rather than the whole search area.
%
Nevertheless, such slight prediction biases have little impact on our BoxCloud comparison and the following tracking process, since they are filtered out by the k-NN grouping.

	\begin{figure*}[t]
	\center\includegraphics[width=0.97\linewidth]{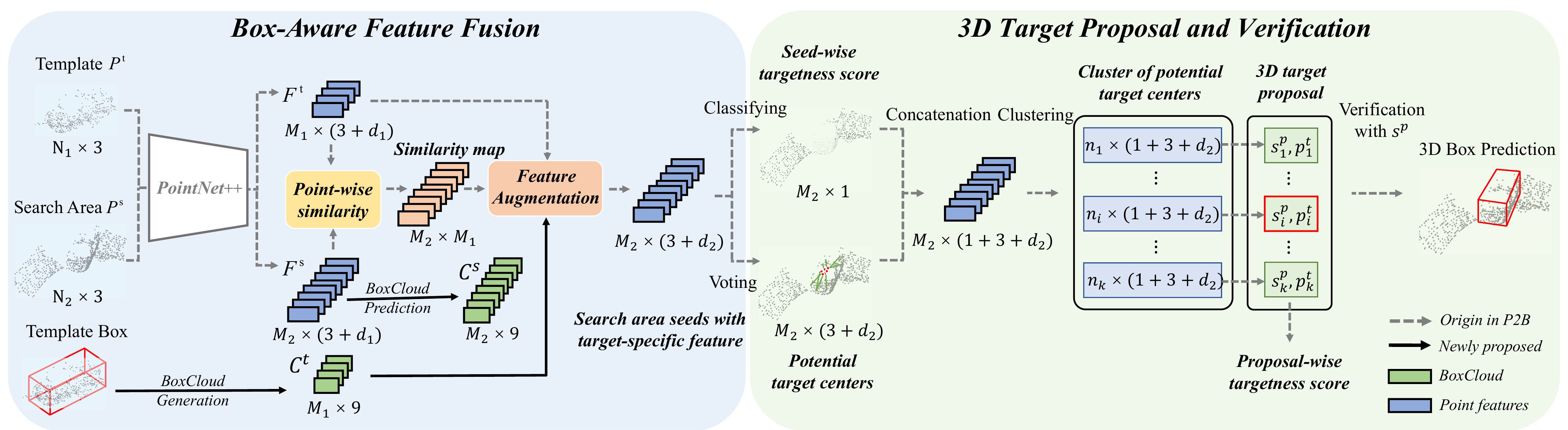}
	\caption{
    \textbf{The overall pipeline of vanilla Box-Aware Tracker (BAT-Vanilla).}  The left part is the box-aware feature fusion, which augments the search area with template information. The right part is a VoteNet-based RPN which generates final target proposals from the target-specific search area. For the $i$-th proposal, $s_i^p$ is its targetness score and $p_i^t$ is its $(x,y,z,\theta)$.
    }
	\label{pipeline}
\end{figure*}
\begin{figure*}[t]
	\center\includegraphics[width=0.97\linewidth]{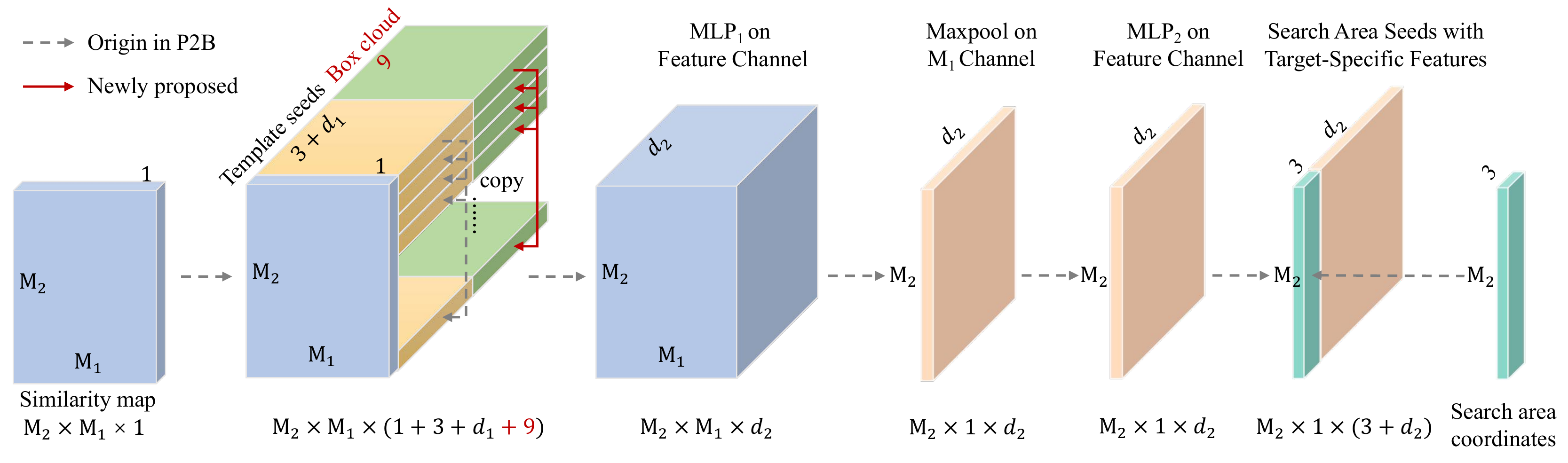}
	\caption{
    \textbf{The detailed architecture of feature augmentation module in the vanilla Box-Aware Tracker (BAT-Vanilla).}  The BoxCloud of the template is concatenated together with the corresponding coordinates and the extracted features.
    }
	\label{augmentation}
\end{figure*}

\section{Architecture of BAT-Vanilla}\label{sec:vanilla}
The Figure~\ref{pipeline} illustrates the architecture of BAT-Vanilla, which directly adds BoxCloud to P2B. After the feature extraction using a shared backbone, a shared MLP (256,256,9) (with layer output sizes 256, 256, 9) is applied to the extracted search area features $F^s$ for BoxCloud prediction. By adding this branch, the $F^s$ is supervised to be box-aware. During the feature augmentation stage, the BoxCloud of the template is simply concatenated together with its coordinates and the extracted features (as shown in Figure~\ref{augmentation}). The feature augmentation in BAT-Vanilla is almost the same with that in P2B, but introduces additional BoxCloud features as priors.

The right part of Figure~\ref{pipeline} illustrates the workflow of the RPN, which is used in both BAT and BAT-Vanilla. A point-wise MLP (256, 256, 3+256) is applied to the target-specific search area features (only the sampled seed points) for object center voting. The MLP takes the per-point feature as input and outputs its offset to the corresponding object center. Besides the coordinate offset (3D), the MLP also predicts a feature offset for each point. But only the coordinate offsets are supervised with the ground-truths. 
In addition to the voting MLP, another MLP (256, 256, 1) is used to predict a targetness score for each search area seed.

After that, the predicted vote centers and targetness scores are concatenated together and then clustered into $k$ groups through the furthest point sampling and the ball query. Finally, a mini-PointNet is used to produce the final target proposal of each group.

\end{document}